\documentclass[sn-nature]{sn-jnl}

\usepackage{graphicx}%
\usepackage{multirow}%
\usepackage{amsmath,amssymb,amsfonts}%
\usepackage{amsthm}%
\usepackage{mathrsfs}%
\usepackage[title]{appendix}%
\usepackage{xcolor}%
\usepackage{textcomp}%
\usepackage{manyfoot}%
\usepackage{booktabs}%
\usepackage{algorithm}%
\usepackage{algorithmicx}%
\usepackage{algpseudocode}%
\usepackage{listings}%

\begin{document}

\title{Collinear datasets augmentation using Procrustes validation sets}

\author*[1]{\fnm{Sergey} \sur{Kucheryavskiy}}\email{svk@bio.aau.dk}

\author[2]{\fnm{Sergei} \sur{Zhilin}}\email{szhilin@gmail.com}

\affil*[1]{\orgdiv{Department of Chemistry and Bioscience}, \orgname{Aalborg University}, \orgaddress{\street{Niels Bohrs vej 8}, \city{Esbjerg}, \postcode{6700}, \country{Denmark}}}

\affil[2]{\orgname{CSort, LLC.}, \orgaddress{\street{Germana Titova st. 7}, \city{Barnaul}, \postcode{656023}, \country{Russia}}}

\abstract{In this paper, we propose a new method for the augmentation of numeric
and mixed datasets. The method generates additional data points by
utilizing cross-validation resampling and latent variable modeling. It
is particularly efficient for datasets with moderate to high degrees of
collinearity, as it directly utilizes this property for generation. The
method is simple, fast, and has very few parameters, which, as shown in
the paper, do not require specific tuning. It has been tested on several
real datasets; here, we report detailed results for two cases,
prediction of protein in minced meat based on near infrared spectra
(fully numeric data with high degree of collinearity) and discrimination
of patients referred for coronary angiography (mixed data, with both
numeric and categorical variables, and moderate collinearity). In both
cases, artificial neural networks were employed for developing the
regression and the discrimination models. The results show a clear
improvement in the performance of the models; thus for the prediction of
meat protein, fitting the model to the augmented data resulted in a
reduction in the root mean squared error computed for the independent
test set by 1.5 to 3 times.}


\maketitle

\section{Introduction}\label{sec1}

Modern machine learning methods that rely on high complexity models,
such as artificial neural networks (ANN), require a large amount of data
to train and optimize the models. Insufficient training data often lead
to overfitting problems, as the number of model hyperparameters to tune
is much larger than the number of degrees of freedom in the dataset.

Another common issue in this case is the lack of reproducibility because
the ANN training procedure is not deterministic, given the random
selection of initial model parameters and the stochastic nature of their
optimization. Consequently, it never leads to a model with the same
parameters and performance, as different training trials can result in
different models. This variability becomes large if the training set is
too small.

This problem is particularly urgent in the case of fitting the
experimental data, as it is often expensive and time-consuming to run
many experimental trials, making it simply impossible to collect
thousands of measurements needed for proper training and optimization.
There can also be other obstacles, such as paperwork related to
permissions in medical research.

One way to overcome the problem of insufficient training data is to
artificially augment it by either simulating new data points or making
small modifications to existing ones. This technique is often referred
to as ``data augmentation''. Data augmentation has proved to be
particularly efficient in image analysis and classification, with a
large body of research reporting both versatile augmentation
methods  \cite{ref_aug_img1} \cite{ref_aug_img2}, \cite{ref_aug_img3}
and methods that are particularly effective for specific cases
\cite{ref_aug_spec1} \cite{ref_aug_spec2}. Augmentation methods
for time series data are also relatively well
developed \cite{ref_aug_time1}.

However, there is a lack of efficient methods that can provide decent
data augmentation for numeric datasets with a moderate to high degree of
collinearity. Such datasets are widespread in experimental research,
including various types of spectroscopic data, results of genome
sequencing (e.g., 16S RNA), and many others. Many tabulated datasets also exhibit internal structures where variables are mutually
correlated. Currently available methods for augmentation of such data
mostly rely on adding various forms of noise \cite{ref_aug_noise} to the
existing measurements, which is not always sufficient. There are also
promising methods that utilize variational autoencoders by random
sampling from their latent variable space \cite{ref_aug_va}, or methods based on
generative adversarial networks \cite{ref_aug_spec1}. The downsides are that both approaches
require building and tuning a specific neural network model for the data
augmentation and hence need a thorough and resource demanding
optimization process and a relatively large initial training set.

In this paper, we propose a simple, fast, versatile, yet efficient
method for augmenting numeric and mixed collinear datasets. The
method is based on an approach that was initially developed for other purposes,
specifically for generating validation sets, and hence is known as Procrustes
cross-validation \cite{ref_pcv1} \cite{ref_pcv2}. However, as demonstrated
in this paper, it effectively addresses the data augmentation problem, resulting
in models with significantly improved prediction or classification
performance.

Our method directly leverages collinearity in the generation
procedure. It fits the training data with a set of latent variables and
then employs cross-validation resampling to measure variations in the
orientation of the variables. This variation is then introduced to the
training set as sampling error, resulting in a new set of data points.

Two fitting models can be employed --- singular value decomposition (SVD) and
partial least squares (PLS) decomposition. The choice of the fitting model
allows the user to prioritize a part of covariance structure, which will
be used for generation of the new data.

Both fitting models have two parameters --- the number of latent
variables and the number of segments used for cross-validation
resampling. The experiments show though that the parameters do not
require specific tuning. Any number of latent variables large enough to
capture the systematic variation of the training set values serve
equally well. As well as any number of segments starting from three.

The proposed method is versatile and can be applied to both fully
numeric data as well as to tabulated data where one or several variables
are qualitative. This opens another perspective, namely data mocking,
which can be useful, e.g., for testing of high loaded software systems,
although we do not consider this aspect here.

The paper describes the theoretical foundations of the method and illustrates its practical application and performance based on two datasets of different nature. It provides comprehensive details on how the method can be effectively applied to diverse datasets in real-world scenarios.

We have implemented the method in several programming languages, including
Python, R, MATLAB, and JavaScript, and all implementations are freely
available in the GitHub repository
(\url{https://github.com/svkucheryavski/pcv}). Additionally, we provide an
online version where one can generate new data points directly in a
browser (\url{https://mda.tools/pcv}).

\section{Methods}\label{sec2}

This chapter presents the general idea behind the Procrustes
cross-validation approach, provides the necessary mathematical
background and describes two implementations of the approach in detail.

Let \(\{\mathbf{X}, \mathbf{Y}\}\) be two matrices of size \(I\times J\)
and \(I\times M\) correspondingly, representing the original training
set to be augmented. Columns of matrix \(\mathbf{X}\) are predictors ---
variables measured or observed directly, for example, light absorbance
at different wavelengths in the case of spectra, abundance of
operational taxonomic units (OTU) in the case of 16S RNA sequencing,
results of clinical tests for patients etc. If the original dataset has
qualitative predictors, they should be replaced by corresponding dummy
variables, which are combined with quantitative predictors to form
\(\mathbf{X}\).

Columns of matrix \(\mathbf{Y}\) are responses ---~variables whose
values should be predicted. There are situations when matrix
\(\mathbf{Y}\) is obsolete, for example, in the case of developing of
one class classifier, only matrix with predictors is needed.

To augment the data a specific model should be developed. It aims at
capturing the variance-covariance structure of \(\mathbf{X}\), or its
part, directly related to the variance of values in \(\mathbf{Y}\). This
model will be referred to as the \emph{PV-model} and denoted as
\(\mathcal{M}\).

The selection of the methods for creating the PV-model depends on the
objectives. Thus, if data augmentation is needed for developing a one
class classifier, when the matrix with responses, \(\mathbf{Y}\), is
obsolete, we suggest using singular value decomposition (SVD), which is
efficient in capturing the whole variance-covariance structure of
\({\mathbf{X}}\).

If the augmented data are used for developing of regression models,
partial least squares (PLS) decomposition, which accounts for the
variance-covariance structure of \(\mathbf{X}^\textrm{T}\mathbf{Y}\)
will be the favorable choice.

In the case of augmenting data for discrimination models, there can be
two solutions. One solution is to consider each class as an independent
population and use SVD decomposition to capture the variance-covariance
structure of \(\mathbf{X}\) within each class. The second solution will
be to create a matrix with dummy predictors, \(\mathbf{Y}\), where each
column corresponds to a particular class, and employ PLS decomposition.
More details about using SVD and PLS decompositions are provided in the
next two sections.

A general definition of the approach can be formulated as follows.

Let us create a \emph{global} PV-model \(\mathcal{M}\) by fitting all
data points from the training set \(\{\mathbf{X}, \mathbf{Y}\}\). If
this model is then applied to a new dataset,
\(\{\mathbf{X}^*, \mathbf{Y}^*\}\), it will result in a set of outcomes
\(\mathbf{R}^*\):

\begin{equation}{
\mathcal{M}(\mathbf{X}^*, \mathbf{Y}^*) \rightarrow \mathbf{R}^*
}\label{eq-pv-outcomes}\end{equation}

The nature of the outcomes depends on the method used for developing of
\(\mathcal{M}\), and can include, for example, predicted response values
in the case of PLS decomposition and scores and distances in the case of
SVD decomposition.

Let us consider cross-validation resampling with \(K\) segments. For any
segment \(k = 1, ..., K\), the dataset \(\{\mathbf{X}, \mathbf{Y}\}\) is
split into a local validation set \(\{\mathbf{X}_k, \mathbf{Y}_k\}\)
with \(I_k = I / K\) rows and a local training set,
\(\{\tilde{\mathbf{X}}_k, \tilde{\mathbf{Y}}_k\}\) with \(I - I_k\)
rows. The local training set is used to create a local PV-model
\(\mathcal{M}_k\), which is then applied to the local validation set to
obtain the outcomes \(\mathbf{R}_k\):

\begin{equation}{
\mathcal{M}_k(\mathbf{X}_k, \mathbf{Y}_k) \rightarrow \mathbf{R}_k
}\label{eq-pv-outcomes-local}\end{equation}

Now let us consider another matrix with predictors, \(\mathbf{X}_{pv}\),
which we will call a \emph{PV-set}. It has the same size as
\(\mathbf{X}\) and the rows of the two matrices are related.

This matrix should be created to obtain the outcomes, computed by
applying the global model \(\mathcal{M}\) to the same \(I_k\) rows of
the PV-set, \(\mathbf{X}_{pv_k}\), to be as close as possible to the
outcomes computed by applying the local model \(\mathcal{M}_k\) to the
local validation set:

\begin{equation}{
\mathcal{M}(\mathbf{X}_{pv_k}, \mathbf{Y}_{k}) \rightarrow \mathbf{R}_{pv_k}
}\label{eq-pv-outcomes-pv}\end{equation}

\begin{equation}{
\mathbf{R}_{pv_k} \approx \mathbf{R}_k
}\label{eq-pv-rule}\end{equation}
for any \(k = 1,...,K\).

This criterion is called a \emph{Procrustean rule}. The main idea of the
proposed approach is that the PV-set, \(\mathbf{X}_{pv}\), created using
the general procedure described above, on the one hand will have a
variance-covariance structure (full or partial), similar to the
variance-covariance structure of the predictors from the original
training set, \(\mathbf{X}\). On the other hand, it will have unique
values, as if it represents another sample taken from the same population
as \(\mathbf{X}\).

Using cross-validation with random splits enables the generation of a
very large number of the PV-sets, which will hold the properties
described above, but will be different from each other. Combining the
generated PV-sets with the original training data will result in the
augmented training set we are aiming at.

The next two sections describe the details of PV-set generation using
SVD and PLS decompositions. More information about the approach can be
found in \cite{ref_pcv2}.

\subsection{Generation of PV-sets based on Singular Value
Decomposition}\label{generation-of-pv-sets-based-on-singular-value-decomposition}

In the case of truncated SVD with \(A\) latent variables, the PV-model
\(\mathcal{M}\) consists of:

\begin{itemize}
\item
  number of latent variables (singular vectors), \(A\)
\item
  \(1\times J\) row-vector for mean centering, \(\mathbf{m}\)
\item
  \(1\times J\) row-vector for standardization (if required),
  \(\mathbf{s}\)
\item
  \(J \times A\) matrix with right singular vectors, \(\mathbf{V}\)
\item
  \(1\times A\) vector with corresponding singular values,
  \(\mathbf{\sigma}\).
\end{itemize}

The SVD decomposition of \(\mathbf{X}\) is defined as:

\begin{equation}{
\mathbf{X} = \mathbf{U}\mathbf{\Sigma}\mathbf{V}^\textrm{T} + \mathbf{E}
}\label{eq-svd1}\end{equation}

Here, \(\mathbf{U}\) is an \(I\times A\) matrix with left singular
vectors and \(\mathbf{\Sigma}\) is an \(A \times A\) diagonal matrix with
singular values \(\mathbf{\sigma} = \{\sigma_1,...,\sigma_A\}\). Element
\(\mathbf{E}\) is an \(I\times J\) matrix with residuals, which can be
used to estimate the lack of fit both for individual data points and for
the whole dataset. For the sake of simplicity we assume that columns of
\(\mathbf{X}\) are already mean centered and standardized.

The right singular vectors \(\mathbf{V}\) act as the orthonormal basis
of the \(A\)-dimensional latent variable space located inside the
original \(J\)-dimensional variable space (\(A \leq J\)). The product of
the left singular vectors and the singular values,
\(\mathbf{T} = \mathbf{U}\mathbf{\Sigma}\), forms \emph{scores} ---
coordinates of data points from \(\mathbf{X}\) being projected to the
latent variable space.

The expression can be also be written as:

\begin{equation}{
\mathbf{X} = \mathbf{TV}^\textrm{T} + \mathbf{E} = \hat{\mathbf{X}} + \mathbf{E}
}\label{eq-svd2}\end{equation}
where \(\hat{\mathbf{X}}\) is a part of \(\mathbf{X}\) that is captured
(or explained) by the SVD model. In the case of full decomposition, when
\(A = rank(\mathbf{X})\), \(\mathbf{X} = \hat{\mathbf{X}}\).

Any new data point, \(\mathbf{x} = \{x_1,...,x_J\}\) can be projected to
the model \(\mathcal{M}\) by mean centering and (if required)
standardizing its values using \(\mathbf{m}\) and \(\mathbf{s}\), and
then projecting the resulting values to the latent variable space
defined by the columns of \(\mathbf{V}\):

\begin{equation}{
\mathbf{t} = \mathbf{x}\mathbf{V}
}\label{eq-svd-score1}\end{equation}

The score vector \(\mathbf{t}\) can be normalized to obtain the
corresponding left singular vector, \(\mathbf{u}\):

\begin{equation}{
\mathbf{u} = \mathbf{t} \mathbf{\Sigma}^{-1} = \{t_1/\sigma_1,...,t_A/\sigma_A\}
}\label{eq-svd-score2}\end{equation}

The explained part of \(\mathbf{x}\) can be computed as:

\begin{equation}{
\hat{\mathbf{x}} = \mathbf{t}\mathbf{V}^T = \mathbf{u}\mathbf{\Sigma}\mathbf{V}^T
}\label{eq-svd-xfull}\end{equation}

The residual part is:

\begin{equation}{
\mathbf{e} = \mathbf{x} - \hat{\mathbf{x}} = \mathbf{x} - \mathbf{u}\mathbf{\Sigma}\mathbf{V} = \mathbf{x}(\mathbf{I} - \mathbf{VV}^\textrm{T})
}\label{eq-svd-xres}\end{equation}

One can define a relationship between the data point \(\mathbf{x}\) and
the PV-model \(\mathcal{M}\) using two distances. A squared Euclidean
distance, \(q\), between \(\mathbf{x}\) and \(\hat{\mathbf{x}}\) in the
original variable space:

\begin{equation}{
q = \sum_{j = 1}^{J} (x_j - \hat{x}_j)^2 = \sum_{j = 1}^{J} e_j^2
}\label{eq-svd-xq}\end{equation}

And a squared Mahalanobis distance, \(h\), between the projection of the
point in the latent variable space and the origin:

\begin{equation}{
h = \sum_{a = 1}^{A} u_a^2
}\label{eq-svd-xh}\end{equation}

The first distance, \(q\), is a measure of lack of fit for the point,
while the second distance, \(h\), is a measure of extremeness of the
point (as the majority of the data points will be located around the
origin due to mean centering).

If we apply cross-validation resampling to the SVD decomposition, it
will result in a set of \(K\) local PV-models, \(\mathcal{M}_k\),
created using the local training sets, \(\tilde{\mathbf{X}}_k\)
(\(k = 1,...K\)).

The main difference between the global PV-model, \(\mathcal{M}\), and
each local PV-model, \(\mathcal{M}_k\), is the orientation of the right
singular vectors, columns of \(\mathbf{V}\) and \(\mathbf{V}_k\), in the
variable space. If we take two singular vectors, the \(a\)-th vector
from the global model, \(\mathbf{v}_a\), and the \(a\)-th vector from
the \(k\)-th local model, \(\mathbf{v}_{ak}\), then the angle between
the vectors in the original variable space can be considered as a
measure of sampling error, emulated by the cross-validation resampling.

If we introduce this error to the local validation set \(\mathbf{X}_k\),
we will create a new set of measurements, \(\mathbf{X}_{pv_{k}}\). In
the case of full rank decomposition this can be done as:

\begin{equation}{
\mathbf{X}_{pv_{k}} = \mathbf{X}_k \mathbf{V}_k \mathbf{V}^\textrm{T}
}\label{eq-svd-xpvk1}\end{equation}
because the dot product \(\mathbf{V}_k \mathbf{V}^\textrm{T}\) is a
rotational matrix between the two orthonormal spaces, defined by
\(\mathbf{V}_k\) and \(\mathbf{V}\).

It can be shown easily that in this case the following Procrustean rules
will be hold:

\begin{equation}{
\mathbf{q}_{pv_{k}} = \mathbf{q}_k
}\label{eq-svd-qrule}\end{equation}

\begin{equation}{
\mathbf{h}_{pv_{k}} = \mathbf{h}_k
}\label{eq-svd-hrule}\end{equation}
where \(\mathbf{q}_{pv_{k}}\) and \(\mathbf{h}_{pv_{k}}\) are vectors
with the two distances, \(q\) and \(h\), computed for each data point
from the \(\mathbf{X}_{pv_{k}}\) being projected to the global model
\(\mathcal{M}\), while \(\mathbf{q_k}\) and \(\mathbf{h_k}\) are vectors
with the two distances computed for each data point from the
\(\mathbf{X}_k\) being projected to the local model \(\mathcal{M}_k\).

Hence we meet the general Procrustes rule requirement defined by
Equation~\ref{eq-pv-rule} with the outcomes \(\mathbf{R}\) consisting of
the two distances. Moreover, this rule holds true for the distances
computed using any number of latent variables \(a = 1,...,A\).

The generation of the PV-set in the case when \(A < rank(\mathbf{X})\)
is similar but requires an additional step to hold the rule defined by
Equation~\ref{eq-svd-qrule}, further details can be found
in \cite{ref_pcv2}.

By repeating this procedure for all segments, \(k = 1,...,K\), we form
the complete PV-set, \(\mathbf{X}_{pv}\), which will have the same
number of data points as the original training set \(\mathbf{X}\) and
will hold the same variance-covariance structure; however it will have
its own sampling error estimated by the cross-validation resampling.

Repeating the generation using random cross-validation splits provides a
large number of the unique PV-sets that can be used to augment the
original training set. Therefore, the SVD based data augmentation
procedure has two parameters:

\begin{itemize}
\item
  number of segments, \(K\)
\item
  number of latent variables, \(A\)
\end{itemize}

As it will be shown in the results section, neither parameter
significantly influences the quality of the augmented data.
In the case of the number of latent variables, \(A\), any number large enough
for capturing the systematic variation in \(\mathbf{X}\), will work
well. The SVD based PV-set generation does not suffer from overfitting,
and optimization of \(A\) is not needed.

\hypertarget{generation-of-pv-sets-based-on-pls-decomposition}{%
\subsection{Generation of PV-sets based on PLS
decomposition}\label{generation-of-pv-sets-based-on-pls-decomposition}}

Partial least squares is a decomposition used for solving multiple
linear regression problem in case when predictors (and responses if they
are multiple) are collinear. The decomposition can be expressed in the
following set of equations:

\begin{equation}{
\mathbf{T} = \mathbf{X} \mathbf{W}
}\label{eq-pls-xdecomp1}\end{equation}

\begin{equation}{
\mathbf{X} = \mathbf{T} \mathbf{P}^\textrm{T} + \mathbf{E}_x = \hat{\mathbf{X}} + \mathbf{E}_x
}\label{eq-pls-xdecomp2}\end{equation}

\begin{equation}{
\mathbf{Y} = \mathbf{T} \mathbf{C}^\textrm{T} + \mathbf{E}_y = \hat{\mathbf{Y}} + \mathbf{E}_y
}\label{eq-pls-ydecomp1}\end{equation}

Here, matrix \(\mathbf{W}\) is a matrix of PLS-weights, which are
computed based on covariance matrix \(\mathbf{X}^\textrm{T}\mathbf{Y}\)
and the consequent deflation procedure using the SIMPLS
algorithm \cite{ref_simpls}. Every column of \(\mathbf{W}\) is a unit
vector, defining the orientation of latent variables in the predictors
space, similar to the columns of \(\mathbf{V}\) in SVD decomposition.
Therefore, the matrix of PLS-scores, \(\mathbf{T}\), contains
coordinates of the original data points being projected to the latent
variables defined by the columns of \(\mathbf{W}\). The number of latent
variables, \(A\), is a decomposition parameter.

In the case of one response, matrix \(\mathbf{Y}\) can be replaced by a
column vector \(\mathbf{y}\) containing \(I\) response values. Then the
Equation~\ref{eq-pls-ydecomp1} can be rewritten in the following compact
form:

\begin{equation}{
\hat{\mathbf{y}} = \mathbf{T} \mathbf{c}^\textrm{T} = c_1 \mathbf{t}_1 + c_2 \mathbf{t}_2 + ... + c_A \mathbf{t}_A
}\label{eq-pls-ydecomp2}\end{equation}

Matrix \(\mathbf{C}\) in this case is represented by a column vector
\(\mathbf{c} = \{c_1, ..., c_A\}\) and vectors
\(\mathbf{t}_1\),\ldots,\(\mathbf{t}_A\) are columns of the score matrix
\(\mathbf{T}\).

In contrast to SVD, the latent variables in PLS are oriented along
directions in the column space of \(\mathbf{X}\), which give the largest
covariance between the scores (columns of matrix \(\mathbf{T}\)) and the
values of \(\mathbf{y}\). Hence, by using PLS, we can prioritize the
variance-covariance structure of \(\mathbf{X}\) directly related to
\(\mathbf{y}\).

The difference between the original and estimated response values can be
defined as:

\begin{equation}{
\mathbf{e} = \mathbf{y} - \hat{\mathbf{y}} = \mathbf{y} - \mathbf{T}\mathbf{c}^\textrm{T}
}\label{eq-pls-ydecomp3}\end{equation}

If we apply cross-validation resampling, then for each segment \(k\)
(\(k = 1,...,K\)), a local PLS-model, \(\mathcal{M}_k\) is computed. The
model consists of, among others, the matrix with weights,
\(\mathbf{W}_k\), and the vector with y-loadings, \(\mathbf{c}_k\),
obtained using the local training set
\(\{\tilde{\mathbf{X}}_k, \tilde{\mathbf{y}}_k\}\).

Applying the local model to the local validation set
\(\{\mathbf{X}_k, \mathbf{y}_k\}\) results in the following:

\begin{equation}{
\mathbf{T}_k = \mathbf{X}_k \mathbf{W}_k
}\label{eq-pls-cvres1}\end{equation}

\begin{equation}{
\hat{\mathbf{y}}_k = \mathbf{T}_k \mathbf{c}_k
}\label{eq-pls-cvres2}\end{equation}

\begin{equation}{
\mathbf{e}_k = \mathbf{y}_k - \hat{\mathbf{y}}_k = \mathbf{y}_k -  \mathbf{T}_k\mathbf{c}_k^\textrm{T}
}\label{eq-pls-cverr}\end{equation}

Now, let us assume that we have a PV-set,
\(\{\mathbf{X}_{pv}, \mathbf{y}\}\). If we take \(I_k\) rows from the
PV-set, \(\{\mathbf{X}_{pv_k}, \mathbf{y}_k\}\), which correspond to the
rows of the local validation set, \(\{\mathbf{X}_k, \mathbf{y}_k\}\),
and apply the global model to the rows we obtain:

\begin{equation}{
\mathbf{T}_{pv_k} = \mathbf{X}_{pv_k} \mathbf{W}
}\label{eq-pls-pvres1}\end{equation}

\begin{equation}{
\hat{\mathbf{y}}_{pv_k} = \mathbf{T}_{pv_k} \mathbf{c}
}\label{eq-pls-pvres2}\end{equation}

\begin{equation}{
\mathbf{e}_{pv_k} = \mathbf{y}_k - \hat{\mathbf{y}}_{pv_k} = \mathbf{y}_k -  \mathbf{T}_{pv_k}\mathbf{c}^\textrm{T}
}\label{eq-pls-cverr2}\end{equation}

We can define the Procrustean rule for the regression as:

\begin{equation}{
\mathbf{e}_{pv_k} = \mathbf{e}_k
}\label{eq-pls-pr1}\end{equation}

Taking into account the equations above, this equation can be simplified
to:

\begin{equation}{
\hat{\mathbf{y}}_{pv_k} = \hat{\mathbf{y}}_k
}\label{eq-pls-pr2}\end{equation}

or in the case of PLS:

\begin{equation}{
\mathbf{T}_{pv_k} \mathbf{c} = \mathbf{T}_{k} \mathbf{c}_k
}\label{eq-pls-pr3}\end{equation}

which gives us the expression for computing \(\mathbf{T}_{pv_k}\):

\[
\mathbf{T}_{pv_k} = \mathbf{T}_{k} \mathbf{D}_k = \mathbf{X}_k \mathbf{W}_k \mathbf{D}_k
\]

where \(\mathbf{D}_k\) is a diagonal matrix with scalars
\(\{c_{k1}/c_1, ..., c_{kA}/c_A\}\) along the main diagonal. Finally,
taking into account Equation~\ref{eq-pls-xdecomp2},
\(\mathbf{X}_{pv_k}\) can be computed as:

\[
\hat{\mathbf{X}}_{pv_k} = \mathbf{T}_{pv_k} \mathbf{P}^\textrm{T} = \mathbf{T}_{k} \mathbf{D}_k \mathbf{P}^\textrm{T}  = \mathbf{X}_k \mathbf{W}_k \mathbf{D}_k \mathbf{P}^\textrm{T}
\]

In contrast to SVD, when points corresponding to \(\mathbf{X}_{pv}\),
are computed mostly by rotations of the points from \(\mathbf{X}\), in
this case a more advanced procedure, which includes both scaling and
rotations, is employed.

In the case of full decomposition, when \(A = rank(\mathbf{X})\),
\(\mathbf{X}_{pv_k} = \hat{\mathbf{X}}_{pv_k}\). When the decomposition
is not full, an additional step is needed in order to compute the matrix
with residuals, \(\mathbf{E}_x\). This step is similar to the one used
for SVD decomposition and can be found in \cite{ref_pcv2}. The
procedure can also be generalized to the case of multiple response
variables, which is also described in \cite{ref_pcv2}.

By repeating this procedure for all \(k = 1,...,K\), the whole PV-set,
\(\mathbf{X}_{pv}\), is computed. Similar to SVD, cross-validation with
random splits creates large number of PV-sets that can be used to
augment the original matrix with predictors, \(\mathbf{X}\). Note, that
the response values for the augmented part will be identical to the
training response (\(\mathbf{y}_{pv} = \mathbf{y}\)).

Therefore, PV-set generation in the case of PLS has the same two
parameters as the SVD-based procedure:

\begin{itemize}
\item
  number of segments, \(K\)
\item
  number of latent variables, \(A\)
\end{itemize}

It must also be noted that the number of latent variables, \(A\), in case of
PLS must be limited (in contrast to the SVD based algorithm). If too
many latent variables are selected, the variation in scalar values, \(c_k/c\),
can be very large, which leads to noisy PV-sets. This happens because
higher latent variables do not have any covariance with the response variable,
\(y\); hence, \(c_k\) values vary chaotically. It is recommended to keep
all latent variables with absolute values of \(c_k/c\) within the \([0, 2]\)
range.

As it is shown in the experimental part, if this limit is introduced,
the number of latent variables does not have a substantial influence on the
PV-sets quality and does not require specific optimization. Same can be
concluded about the number of segments, \(K\).

\section{Results}\label{sec3}
\subsection{Datasets}\label{datasets}

Two datasets were selected to demonstrate the capabilities of PV-sets to
be used for data augmentation. Since both SVD and PLS decompositions
model the variance-covariance structure of \(\mathbf{X}\), the proposed
augmentation approach will work best for datasets with moderate to high
degree of collinearity, which was taken into account for the selection
of the datasets.

It must be noted, that optimization of the ANN architecture was outside
the scope of this work. In both examples a very simple architecture with
several layers was used.

All calculations were carried out using Python 3.10.12 supplemented with
packages \emph{PyTorch}\cite{ref_pytorch} 2.0.1,
\emph{NumPy}\cite{ref_numpy} 1.26.0 and \emph{prcv} 1.1.0.
Statistical analysis of the results was performed in R 4.3.1.

Python scripts used to obtain the results, presented in this chapter,
are freely available so everyone can reproduce the reported results.

\subsubsection{Tecator}

The \emph{Tecator} dataset consists of spectroscopic measurements of
finely minced meat samples with different moisture, protein and fat
content. The measurements were taken by Tecator Infratec Food and Feed
Analyzer working by the Near Infrared Transmission (NIT) principle. The
dataset was downloaded from the StatLib public dataset archive
(\url{http://lib.stat.cmu.edu/datasets/}).

The dataset contains the spectra and the response values for 215 meat
samples in total (subsets labeled as \emph{C}, \emph{M} and \emph{T} in
the archive), of which 170 samples are used as the training set and 45
are used as the independent test set to assess the performance of the
final optimized models. Fat content (as \% w/w) was selected as the
response variable for this research.

The matrix of predictors \(\mathbf{X}\) contains absorbance values
measured at 100 wavelengths in the range of 850--1050 nm.

\begin{figure}
{\centering \includegraphics[width=1.0\textwidth]{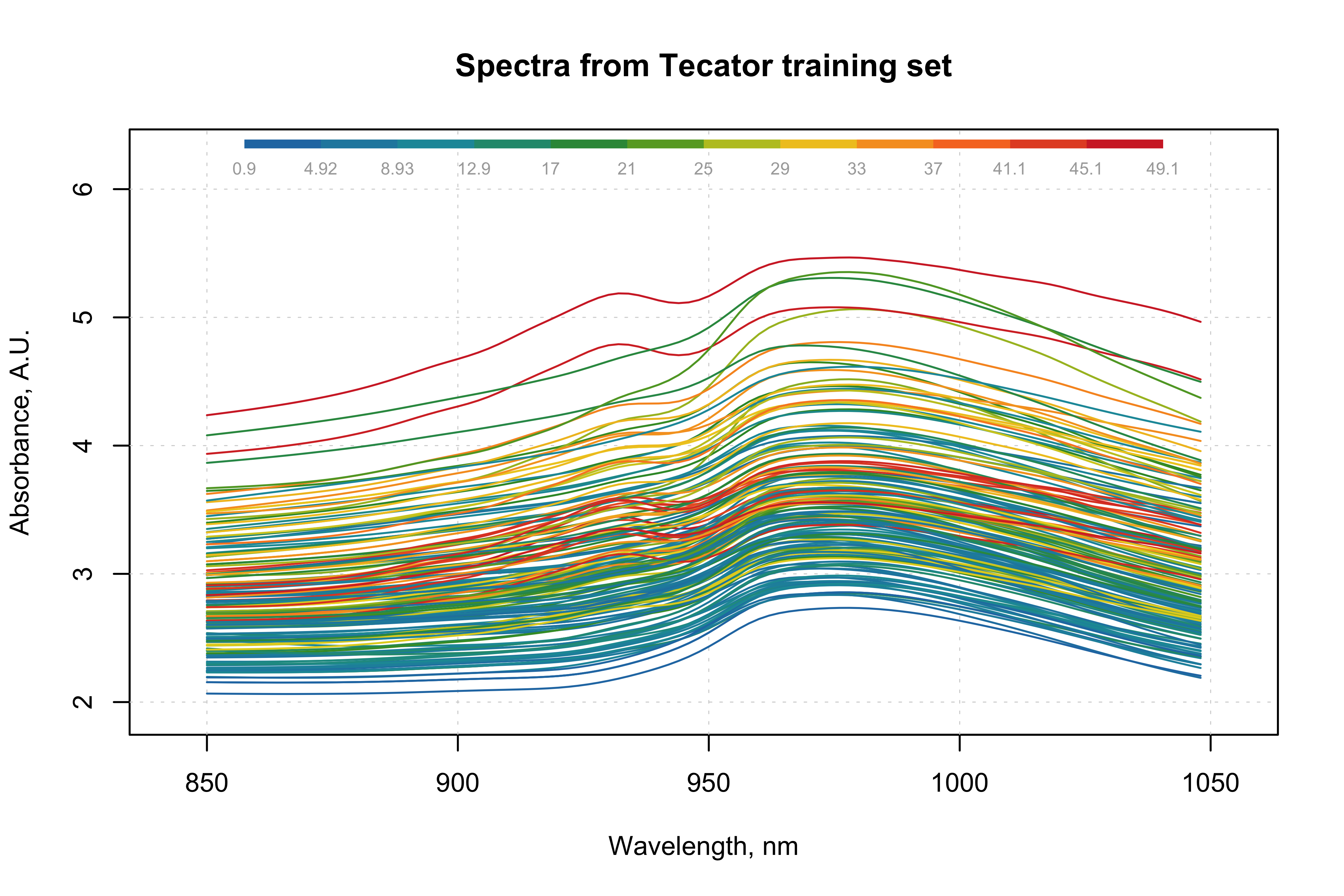}}
\caption{\label{fig-tecator-spectra}Tecator spectra colored by fat
content in the samples (colorbar legend maps the colors of the spectra
to the fat content in \%w/w).}
\end{figure}

According to the Beer-Lambert law, there is a linear relationship
between the concentration of chemical components and the absorbance of
the electromagnetic radiation. Therefore, the prediction of fat content
can in theory be performed by fitting the dataset with a multiple linear
regression model. However, the shape of the real spectra suffers from
various side effects, including noise, light scattering from an uneven
surface of the samples, and many others.

Figure~\ref{fig-tecator-spectra} shows the spectra from the training set
in the form of line plots colored according to the corresponding fat
content using a color gradient from blue (low content) to red (high
content) colors. As one can see, there is no clear association between
the shape of the spectra and the response value.

Handling such data necessitates the careful selection and optimization
of preprocessing methods to eliminate undesirable effects and reveal the
information of interest. However, the use of nonlinear models, such as
artificial neural networks, has the capability to automatically unveil
the needed information \cite{ref_tecator}, bypassing the need for
extensive preprocessing steps.

Nonlinearity is not the sole factor in play. Preprocessing can be
considered as a feature extraction procedure. It is known that in
networks with many hidden layers, the initial layers act as feature
extractors. The greater the number of layers, the more intricate
features can be derived from raw data. However, it is worth noting that
such models demand a substantial number of measurements or observations
to achieve satisfactory performance.

The \emph{Tecator} dataset will be employed for a thorough investigation
of how the use of PV-set based data augmentation can attack the problem
and how the two main parameters of the PV-set generation procedure
(number of latent variables, number of segments) as well as the number of
generated PV-sets influence the performance of the ANN regression
models.

\textbf{Heart disease}

The \emph{Heart} dataset came from a study of 303 patients referred for
coronary angiography at the Cleveland Clinic. The patients were examined
by a set of tests, including physical examination, electrocardiogram at
rest, serum cholesterol determination and fasting blood sugar
determination. The dataset consists of the examination results combined
with historical records. More details about the data can be found
elsewhere \cite{ref_heart1} \cite{ref_heart2}. The dataset is publicly available from
the UC Irvine Machine Learning Repository \cite{ref_heart_data_repo}. The
original data include records from several hospitals, in this research
only data from the Cleveland Clinic are used.

Eleven records with missing values were removed from the original data,
resulting in 292 rows. The dataset consists of 14 numeric and
categorical variables, the overview is given in
Table~\ref{tbl-heart-dataset}. The \emph{Class} variable is used as a
response, and the rest are attributed to predictors.

\begin{table}[h]
\caption{Variables of the \emph{Heart} dataset, their type and value range.}\label{tbl-heart-dataset}
\begin{tabular}{@{}lll@{}}
\toprule
Attribute & Type  & Values\\
\midrule
Age & numeric, years & 29--77 \\
Sex & categorical & male (200), female (92) \\
Chest pain type & categorical & abnang (49), angina (23), asympt (138),
notang (82) \\
Resting blood pressure & numeric & 94--200 \\
Cholesterol & numeric & 126--564 \\
Sugar & categorical & \(<120\) (43), \(\geq120\) (249) \\
Resting ECG & categorical & hyper (145), normal (147) \\
Max heart rate & numeric & 71--202 \\
Exercise induced angina & categorical & true (95), false (197) \\
Oldpeak & numeric & 0--6.2 \\
Slope & categorical & down (20), flat (134), up (138) \\
Number of vessels colored & discrete & 0--3 \\
Thal & categorical & fix (17), normal (161), rev (114) \\
Class & categorical & healthy (159), sick (133) \\
\botrule
\end{tabular}
\end{table}

This dataset is used to demonstrate that PV-set augmentation can also be
applied to mixed datasets with categorical variables, and to show how
SVD and PLS versions work for solving binary classification tasks.

\subsection{ANN regression of Tecator data}\label{ann-regression-of-tecator-data}

As with most of the spectroscopic measurements, the \emph{Tecator} data
are highly collinear, and SVD decomposition of the matrix with spectra
resulted in the first eigenvalue of 25.6, while starting from the 6th
all eigenvalues are significantly below 0.001. PLS decomposition of the
training set resulted in similar outcomes, therefore, one can assume
that using 5-6 latent variables is enough to capture the majority of the
predictors' variation.

At the same time, preliminary investigation has shown that \(c_k/c\)
values are within the desired limit of \([0, 2]\) for all latent variables up
to \(A = 50\). Thus, using any number of latent variables between \(5\) and
\(50\) will produce a reasonable PV-set.

To predict fat content a simple ANN model was employed. The model
consisted of six fully connected (linear) layers of the following sizes:
\((100, 150)\), \((150, 200)\), \((200, 150)\), \((150, 100)\),
\((100, 50)\), and \((50, 1)\). The first value in the parenthesis
denotes the number of inputs and the second value denotes the number of
outputs in each layer. The first five layers were supplemented with the
ReLU activation function, while the last layer was used as the output of
the model. All other characteristics of the model are shown in
Table~\ref{tbl-tecator-model-params}.

The model contains approximately 95000 tunable parameters in total,
which is much larger than the number of samples in the training set
(170). However, as already noted, first several hidden layers of ANN
usually serve as feature extractors, so not all parameters will be a
part of the regression model itself.

\begin{table}[h]
\caption{Characteristics of the ANN model for Tecator experiments.}\label{tbl-tecator-model-params}
\begin{tabular}{@{}ll@{}}
\toprule
Characteristic & Value\\
\midrule
Layer 1 & (100, 150) + ReLU \\
Layer 2 & (150, 200) + ReLU \\
Layer 3 & (200, 150) + ReLU \\
Layer 4 & (150, 100) + ReLU \\
Layer 5 & (100, 50) + ReLU \\
Layer 6 & (50, 1) \\
Optimizer & Adam \\
Learning rate & 0.0001 \\
Number of epochs & 300 \\
Batch size & 10 \\
\botrule
\end{tabular}
\end{table}

The columns of the predictors were standardized using mean and standard
deviation values computed for the original training set. The response
values were mean centered only to obtain errors directly comparable in
the original units (\%w/w).

First, the ANN training procedure was run 30 times using the original
training set without augmentation. Repeated experiments are necessary
because the ANN training procedure is not deterministic, as explained in
the introduction, and the performance varies from model to model.

At each run the resulting model was evaluated using the independent test
set. The median root mean squared error for the test set (RMSEP) was
\(4.17\) \%w/w and the coefficient of determination, \(R^2\), was
\(0.91\). The best values for RMSEP and \(R^2\) were \(3.69\) and
\(0.93\) respectively.

After that, a full factorial experiment was set up to determine how the
augmentation of the training data with PV-sets influences the
performance of the ANN model and which parameters for PV-set generation
as well as the number of generated sets have the largest effect on the
performance. The PV-sets were generated using PLS decomposition. The
generation parameters are shown in Table~\ref{tbl-tecator-pvset-params}.

All possible combinations of the values from the table were tested (60
combinations in total). For every combination the training/test
procedure was repeated 5 times with full reinitialization, resulting in
300 outcomes.

\begin{table}[h]
\caption{Levels of PV-set generation parameters and number of sets used in the factorial experiments.}\label{tbl-tecator-pvset-params}
\begin{tabular}{@{}ll@{}}
\toprule
Parameter & Levels\\
\midrule
Number of PV-sets & 1, 5, 10, 20, 50 \\
Number of LV-s & 5, 10, 20 \\
Number of segments & 2, 4, 10, 20 \\
\botrule
\end{tabular}
\end{table}

\begin{figure}
{\centering \includegraphics[width=1.0\textwidth]{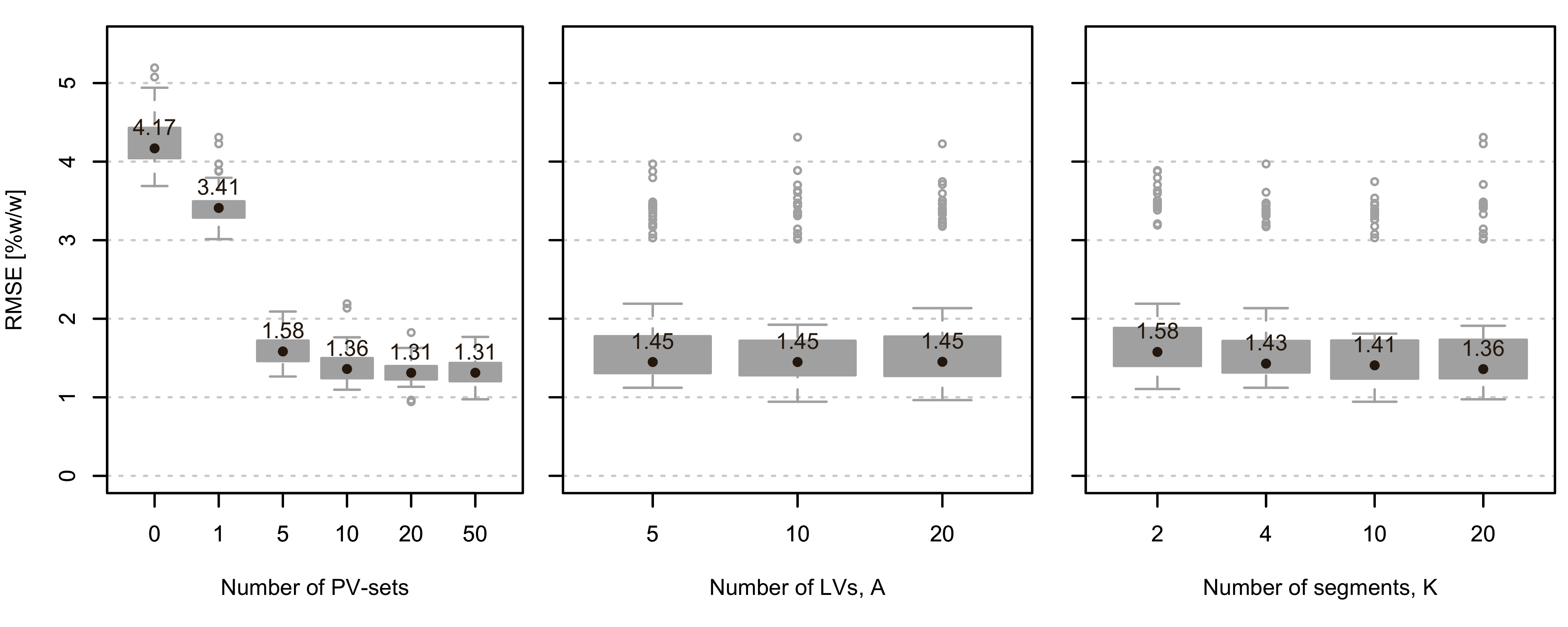}}
\caption{\label{fig-tecator-annres1}Box and whiskers plots showing
variation of RMSEP values of the ANN model depending on the parameters
of PV-sets generation and the number of the sets (left --- number of
PV-sets, middle --- number of latent variables in PV-model, right --- number
of segments in cross-validation resampling). Black points and numbers
show median values.}
\end{figure}

Figure~\ref{fig-tecator-annres1} graphically illustrates the outcomes of
the experiment in the form of boxplots. Every plot shows the variation
in the RMSEP for a given value of one of the three tested parameters.
The black point inside each box supplemented with text shows the
corresponding median value.

It is clear that the number of PV-sets used for data augmentation (first
plot in the figure) has the largest effect on the RMSEP. Thus,
augmenting the original training set with one PV-set reduces the median
test set error by approximately 18\% (from \(4.17\) to \(3.41\) \%w/w).
Adding five PV-sets reduces the median RMSEP down to \(1.58\) \%w/w
(62\% reduction). Using 10 PV-sets further reduces the RMSEP down to
\(1.36\) \%w/w; however, the effect is smaller.

Changing the number of latent variables does not give a significant difference
in the performance of the models, while the number of segments in
cross-validation resampling shows a small effect. The best model was
obtained by using data augmented with 20 PV-sets generated with 10
latent variables and 10 segments. It could predict fat content in the test set
with RMSEP = 0.94 \%w/w (\(R^2 = 0.995\)) which is more than 3 times
smaller compared to the best model obtained without augmentation.

\begin{figure}
{\centering \includegraphics[width=1.0\textwidth]{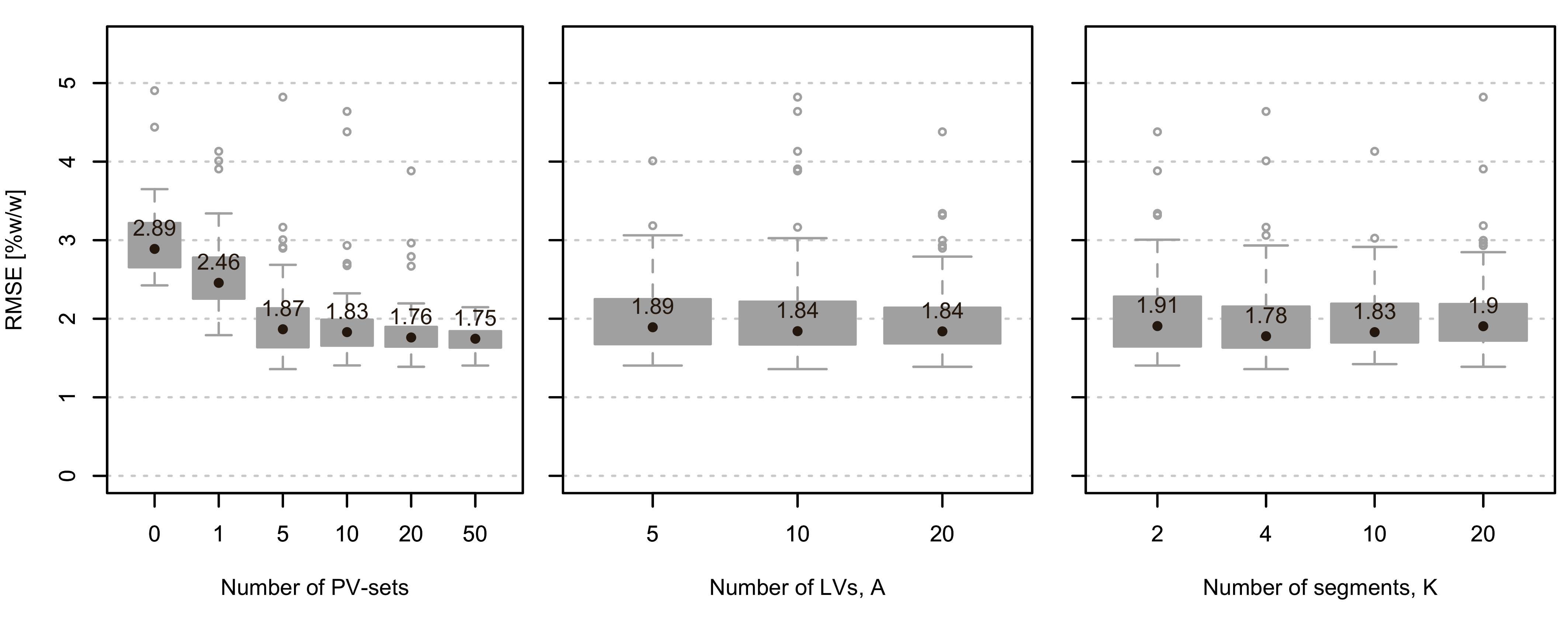}}
\caption{\label{fig-tecator-annres2}Box and whiskers plots showing
variation of RMSEP values of the ANN model depending on the parameters
of PV-sets generation and the number of the generated sets, obtained
using larger learning rate (0.001).}
\end{figure}

Statistical analysis of the outcomes carried out by N-way analysis of
variance (ANOVA) and Tukey's honest significance difference (HSD) test
confirmed the significant difference between average RMSE values
obtained using different numbers of PV-sets (ANOVA p-value \( \ll 0.01\)).
However, pairwise comparison shows no significant difference between
10 vs. 20, 10 vs. 50, and 20 vs 50 PV-sets (p-value \(>0.20\) in for all three pairs).

The number of latent variables has no significant effect (ANOVA p-value \(\approx 0.71\)).
The number of segments shows only a significant difference between 2 segments and other
choices, favoring the use of 4 segments or above (p-values for all pairs with
2 segments \(\ll0.01\), for 4 segments vs. 10 segments p-value \(\approx 0.26\))

Optimization of the ANN learning parameters (by varying the learning
rate and the batch size) toward reducing RMSEP values for the models
trained without augmentation, made this gap smaller.
Figure~\ref{fig-tecator-annres2} shows the results obtained using a
learning rate = 0.001. Training the model without augmentation results
with median RMSEP = \(2.84\) \%w/w. Training the model on the augmented
data with 20 PV-sets resulted in a median RMSEP = \(1.80\) \%w/w. This
is still approximately 37\% smaller; however, the overall performance of
the model trained with this learning rate is worse.

\begin{figure}
{\centering \includegraphics[width=1.0\textwidth]{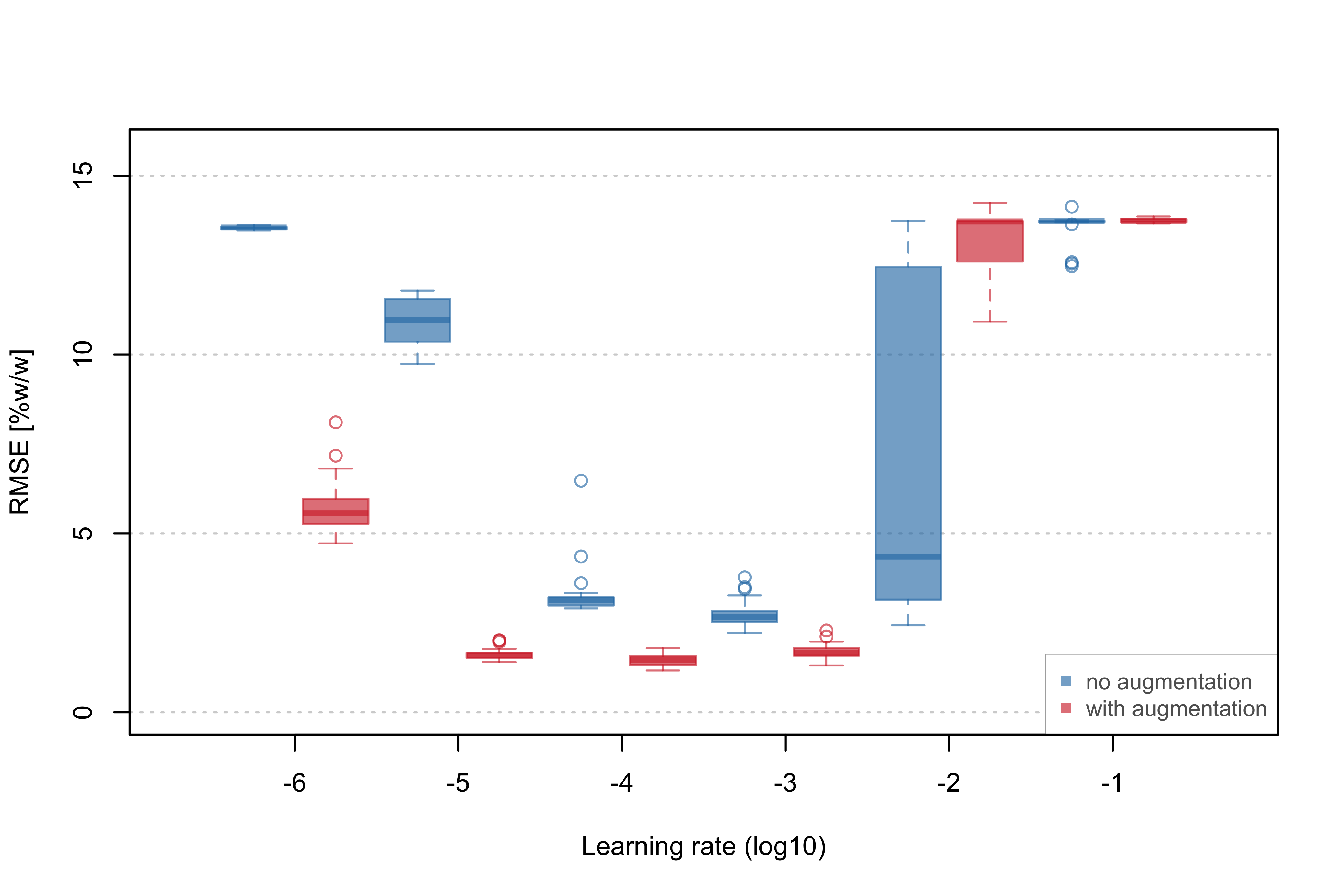}}
\caption{\label{fig-tecator-lr}Box and whiskers plots showing variation
of RMSEP values of the ANN models trained using original training data
and augmented data depending on the learning rate.}
\end{figure}

This effect is understandable as changing the learning rate and batch
size to improve the performance of the model trained on the original
data makes the model less sensitive to overfitting and local minima
problems but also less flexible, which, in turn, makes the use of
augmented data less efficient.

Figure~\ref{fig-tecator-lr} further illustrates this effect. The plot
shows the variation in RMSEP values for ANN models trained using
different learning rates. Each learning rate is represented by a pair of
box and whiskers series. The left (blue) series in each pair illustrates
the results obtained using 30 models trained with the original data. The
right (red) series in each pair shows the results for 30 models trained
using the augmented data (20 PV-sets computed using 10 LVs and 4
segments). New PV-sets were computed at each run to eliminate possible
random effects.

As one can see, at small learning rates (\(10^{-5}\)---\(10^{-3}\)) the
models trained with augmented data clearly outperform the models trained
using the original training set. Not only in mean or median RMSEP values
but also model-to-model variation of RMSEP is much smaller, despite the
additional uncertainty introduced by the augmentation (each PV-set is
generated using random splits; hence, the augmented datasets are always
different from each other). Starting from LR = \(10^{-2}\) and larger
data augmentation does not show any benefits, but the model trained
using the original data also clearly loses the performance.

This experiment was also repeated using other ANN architectures,
including networks with convolution layers. The gap between the
performance varied depending on the architecture and the learning
parameters; however, in all experiments, a minimum of 20\% remained,
clearly indicating the benefits of the PV-set augmentation in this
particular case.

\hypertarget{ann-classification-of-heart-data}{%
\subsection{ANN classification of Heart
data}\label{ann-classification-of-heart-data}}

All categorical variables from the \emph{Heart} dataset with \(L\)
levels were converted to \(L-1\) dummy values \([0, 1]\), so the matrix
with predictors, \(\mathbf{X}\) contained 17 columns in total. The
columns of the matrix were mean centered and standardized. SVD
decomposition of the matrix indicates moderate collinearity with
eigenvalues ranging from 3.43 to 0.77 for the first 10 latent variables.

ANN model for classification of the patient conditions used in this
research included six linear layers and one sigmoid layer. The first
five layers were supplemented with the ReLU activation function, while
the last layer was used as the output of the model. The main
characteristics of the model are shown in
Table~\ref{tbl-heart-model-params}. The model has approximately 14700
tunable parameters, while the original dataset has only 292 objects.

\begin{table}[h]
\caption{Characteristics of the ANN model for Heart dataset classification.}\label{tbl-heart-model-params}
\begin{tabular}{@{}ll@{}}
\toprule
Characteristic & Value \\
\midrule
Layer 1 & (17, 34) + ReLU \\
Layer 2 & (34, 68) + ReLU \\
Layer 3 & (68, 68) + ReLU \\
Layer 4 & (68, 68) + ReLU \\
Layer 5 & (68, 34) + ReLU \\
Layer 6 & (34, 1) \\
Layer 7 & Sigmoid \\
Optimizer & Adam \\
Learning rate & 0.000001 \\
Number of epochs & 300 \\
Batch size & 10 \\
\botrule
\end{tabular}
\end{table}

Since this dataset does not contain a dedicated test set, the following
procedure was employed. At each run, the data objects were split into
two groups with healthy persons in one group and sick persons in the
other. Then, 75\% of records were selected randomly from each group and
merged together to form a training set. The remaining 25\% of records
formed the test set for the run. Classification accuracy, which was
computed as a ratio of all correctly classified records to the total
number of records in the test set, was used as the performance
indicator.

The experimental design was similar to the \emph{Tecator} experiments.
First, the training/test procedure was repeated 30 times using the data
without augmentation. Then the ANN models were trained using the
augmented data and tested using the randomly selected test set. The
PV-sets for augmentation were computed using an algorithm based on PLS
decomposition applied to the randomly selected training set. As in the
previous chapter, all possible combinations of PV-set generation
parameters from Table~\ref{tbl-tecator-model-params} were tested with 5
replications (300 runs in total).

\begin{figure}
{\centering \includegraphics[width=1.0\textwidth]{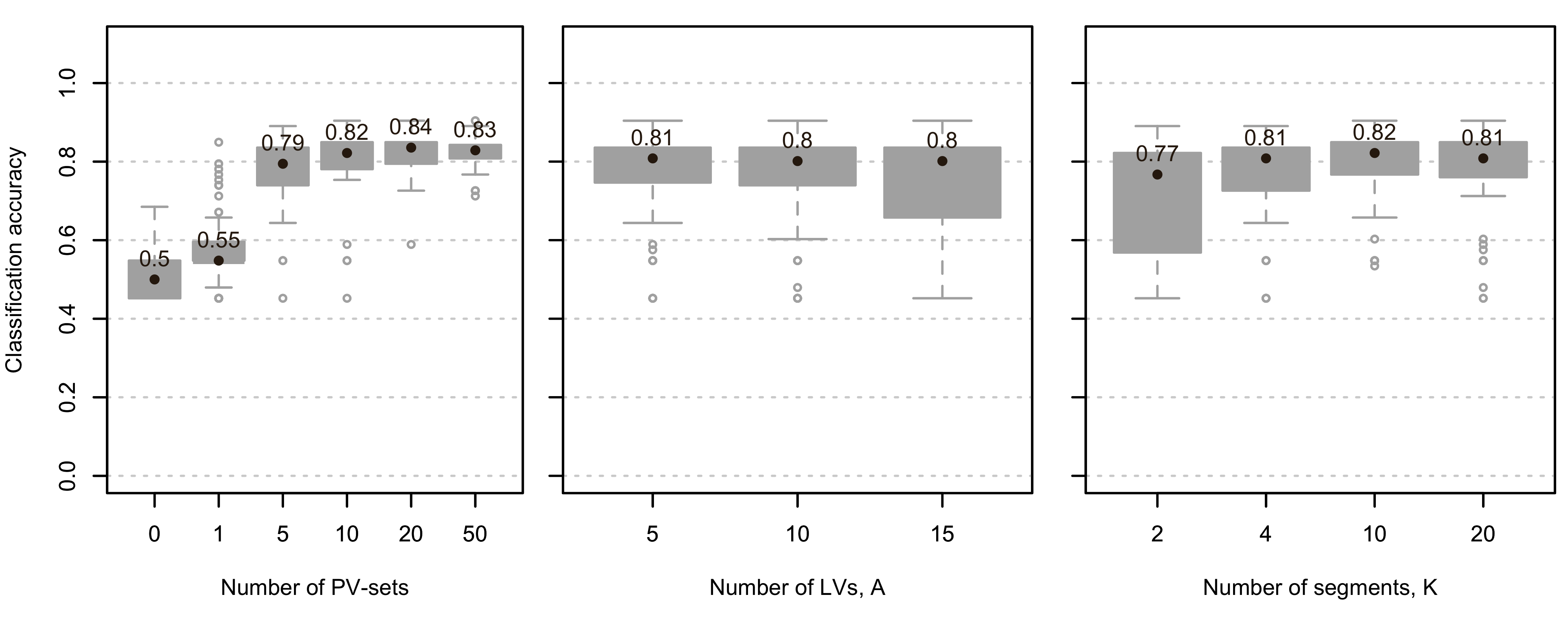}}
\caption{\label{fig-heart-annres1}Box and whiskers plots showing
variation of test set accuracy of the ANN models depending on PV-set
generation parameters (PV-sets computed using algorithm based on PLS
decomposition, ANN learning rate = \(10^{-6}\)).}
\end{figure}

Figure~\ref{fig-heart-annres1} presents the outcomes of the experiment
in the form of boxplots, which show a variation of the test set accuracy
computed at each run. One can clearly notice that the accuracy of the
models trained on the data without augmentation is very low (median
accuracy is 0.50), while augmenting the training data with 20 PV-sets
raises the median accuracy to 0.84. The best model obtained in these
experiments is also trained on the augmented data and has an accuracy of
0.91.

N-way ANOVA and Tukey's tests for the outcomes have shown that two
parameters --- number of PV-sets as well as number of segments in
cross-validation resampling have a significant influence on the accuracy
(p-value for both factors was \(\ll 0.001\) in ANOVA test). However, a
significant difference was observed only for the number of segments
equal to two, similar to the \emph{Tecator} results. Using four or more
segments for the generation of PV-sets shows statistically similar
performance to the models trained on the augmented data.

\begin{figure}
{\centering \includegraphics[width=1.0\textwidth]{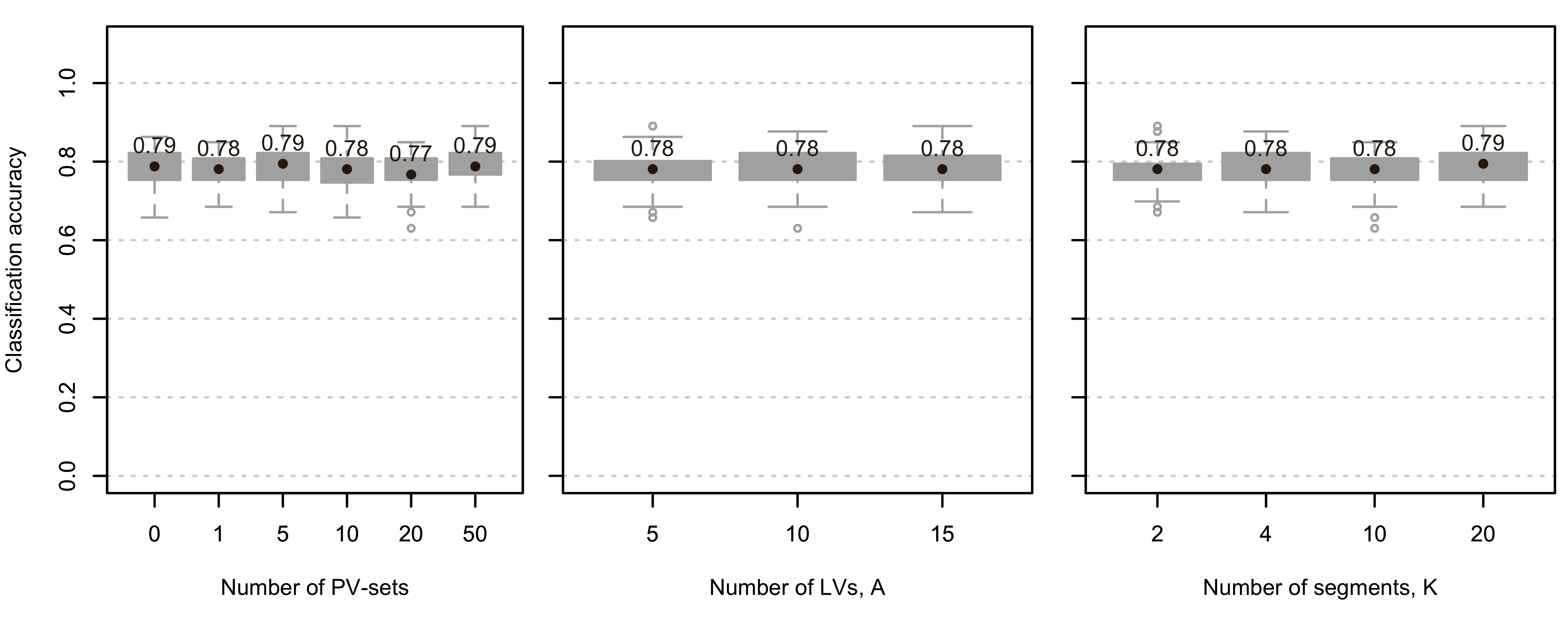}}
\caption{\label{fig-heart-annres2}Box and whiskers plots showing
variation of test set accuracy of the ANN models depending on parameters
of PV-set generation (PV-sets computed using algorithm based on PLS
decomposition, ANN learning rate = \(10^{-3}\)).}
\end{figure}

The large gap in the performance of the models trained using the
original and augmented data is explained by the initial selection of a
low learning rate (\(10^{-6}\)), which was justified by the results of
experiments with \emph{Tecator} data. Optimization of the learning rate
to obtain the best accuracy for the data without augmentation eliminates
this gap as shown in Figure~\ref{fig-heart-annres2}. However, as in the
\emph{Tecator} experiments, the overall performance of the models in
this case is slightly lower, and the median accuracy was approximately
0.79.

\begin{figure}
{\centering \includegraphics[width=1.0\textwidth]{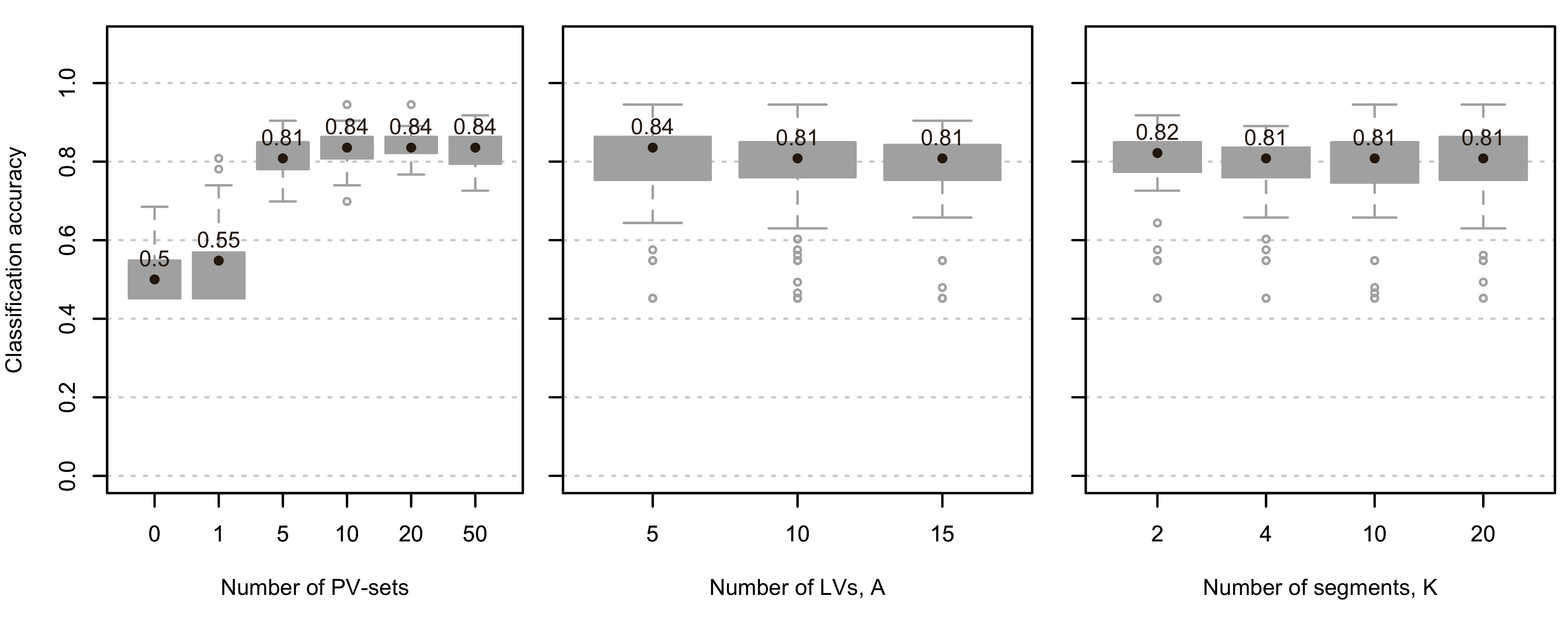}}
\caption{\label{fig-heart-annres3}Box and whiskers plots showing
variation of test set accuracy of the ANN models depending on PV-set
generation (PV-sets computed using algorithm based on SVD decomposition,
ANN learning rate = \(10^{-6}\)).}
\end{figure}

Since the \emph{Heart} data consist of two classes and the classes are
balanced (the number of observations in the classes is comparable), one
can apply an SVD-based algorithm for PV-set generation. In this case, at
each run the PV-sets are generated separately for the sick and healthy
groups, and then are merged together to form the augmented training set.
The results obtained for the learning rate of \(10^{-6}\) are shown in
Figure~\ref{fig-heart-annres3}.

The results are very similar to those obtained using the PLS-based
algorithm. However, in this case, the overall performance of the models
trained on the augmented data was a slightly higher, with a median
accuracy of 0.84 for the models trained with on data augmented with 10,
20 and 50 PV-sets. The best model had an accuracy of 0.95.

Reducing the learning rate down to 0.001 had the same effect as for the
data augmented using the PLS-based algorithm --- the gap between the
models trained with and without augmented data is eliminated however the
overall performance also gets lower with, median a accuracy of
approximately 0.79.

\section{Discussion}\label{sec4}

The experimental results confirm the benefits of PV-set augmentation, however
optimization of ANN learning parameters is needed to make the benefits
significant. At the same time, optimization of the PV-set generation algorithm is
not necessary for most of the cases. Based on our experiments, we advise
using cross-validation resampling with 5 or 10 splits and a number of
latent variables large enough to capture the majority of variation in
\(\mathbf{X}\). In some specific cases one can use tools for quality
control of generated PV-sets described in \cite{ref_pcv2}.

It must also be noted that the use of PV-sets for data augmentation is
not always beneficial. Thus, according to our experiments, which are not
reported here, in methods that are robust to overfitting, such as, for
example, random forest (RF), increasing the training set artificially
does not have a significant effect on the model performance. In the case
of eXtreme Gradient Boosting changing training parameters, which
regulate the overfitting, such as a learning rate, maximum depth and
minimum sum of instance weight, can have an effect, but most of the time
the effect observed in our experiments was marginal.

\section{Conclusions}\label{sec5}

This paper proposes a new method for data augmentation. The method is
beneficial specifically for datasets with moderate to high degree of collinearity
as it directly utilizes this feature in the generation algorithm.

Two proposed implementations of the method (SVD and PLS based) cover most
of the common data analysis tasks, such as regression, discrimination and
one-class classification (authentication). Both implementations are very
fast --- the generation of a PV-set for \(\mathbf{X}\) of \(200 \times 500\) with 20
latent variables and 10 segments splits requires several seconds (less
than a second on a powerful PC), much less than the training of an ANN model
with several layers.

The method can work with datasets of small size (from tens observations)
and can be used for both numeric and mixed datasets, where one or several
variables are categorical.


\begin{thebibliography}{10}
\expandafter\ifx\csname url\endcsname\relax
  \def\url#1{\burl{#1}}\fi
\expandafter\ifx\csname urlprefix\endcsname\relax\def\urlprefix{URL }\fi
\providecommand{\bibinfo}[2]{#2}
\providecommand{\eprint}[2][]{\url{#2}}
\providecommand{\doi}[1]{\url{https://doi.org/#1}}
\bibcommenthead

\bibitem{ref_aug_img1}
\bibinfo{author}{Ratner, A.~J.}, \bibinfo{author}{Ehrenberg, H.~R.}, \bibinfo{author}{Hussain, Z.}, \bibinfo{author}{Dunnmon, J.} \& \bibinfo{author}{Ré, C.}
\newblock \bibinfo{title}{Learning to compose domain-specific transformations for data augmentation} (\bibinfo{year}{2017}).
\newblock \eprint{1709.01643}.

\bibitem{ref_aug_img2}
\bibinfo{author}{Goodfellow, I.~J.} \emph{et~al.}
\newblock \bibinfo{title}{Generative adversarial networks} (\bibinfo{year}{2014}).
\newblock \eprint{1406.2661}.

\bibitem{ref_aug_img3}
\bibinfo{author}{Dao, T.} \emph{et~al.}
\newblock \bibinfo{title}{A kernel theory of modern data augmentation} (\bibinfo{year}{2019}).
\newblock \eprint{1803.06084}.

\bibitem{ref_aug_spec1}
\bibinfo{author}{Pérez, E.} \& \bibinfo{author}{Ventura, S.}
\newblock \bibinfo{title}{Progressive growing of generative adversarial networks for improving data augmentation and skin cancer diagnosis}.
\newblock \emph{\bibinfo{journal}{Artificial Intelligence in Medicine}} \textbf{\bibinfo{volume}{141}}, \bibinfo{pages}{102556} (\bibinfo{year}{2023}).
\newblock \urlprefix\url{https://www.sciencedirect.com/science/article/pii/S0933365723000702}.

\bibitem{ref_aug_spec2}
\bibinfo{author}{Perez, F.}, \bibinfo{author}{Vasconcelos, C.}, \bibinfo{author}{Avila, S.} \& \bibinfo{author}{Valle, E.}
\newblock \bibinfo{editor}{Stoyanov, D.} \emph{et~al.} (eds) \emph{\bibinfo{title}{Data augmentation for skin lesion analysis}}.
\newblock (eds \bibinfo{editor}{Stoyanov, D.} \emph{et~al.}) \emph{\bibinfo{booktitle}{OR 2.0 Context-Aware Operating Theaters, Computer Assisted Robotic Endoscopy, Clinical Image-Based Procedures, and Skin Image Analysis}}, \bibinfo{pages}{303--311} (\bibinfo{publisher}{Springer International Publishing}, \bibinfo{address}{Cham}, \bibinfo{year}{2018}).

\bibitem{ref_aug_time1}
\bibinfo{author}{Iglesias, G.}, \bibinfo{author}{Talavera, E.}, \bibinfo{author}{Gonz{\'a}lez-Prieto, {\'A}.}, \bibinfo{author}{Mozo, A.} \& \bibinfo{author}{G{\'o}mez-Canaval, S.}
\newblock \bibinfo{title}{Data augmentation techniques in time series domain: a survey and taxonomy}.
\newblock \emph{\bibinfo{journal}{Neural Computing and Applications}} \textbf{\bibinfo{volume}{35}}, \bibinfo{pages}{10123--10145} (\bibinfo{year}{2023}).
\newblock \urlprefix\url{https://doi.org/10.1007/s00521-023-08459-3}.

\bibitem{ref_aug_noise}
\bibinfo{author}{Sáiz-Abajo, M.}, \bibinfo{author}{Mevik, B.-H.}, \bibinfo{author}{Segtnan, V.} \& \bibinfo{author}{Næs, T.}
\newblock \bibinfo{title}{Ensemble methods and data augmentation by noise addition applied to the analysis of spectroscopic data}.
\newblock \emph{\bibinfo{journal}{Analytica Chimica Acta}} \textbf{\bibinfo{volume}{533}}, \bibinfo{pages}{147--159} (\bibinfo{year}{2005}).
\newblock \urlprefix\url{https://www.sciencedirect.com/science/article/pii/S000326700401428X}.

\bibitem{ref_aug_va}
\bibinfo{author}{Chadebec, C.} \& \bibinfo{author}{Allassonnière, S.}
\newblock \bibinfo{title}{Data augmentation with variational autoencoders and manifold sampling} (\bibinfo{year}{2021}).
\newblock \eprint{2103.13751}.

\bibitem{ref_pcv1}
\bibinfo{author}{Kucheryavskiy, S.}, \bibinfo{author}{Zhilin, S.}, \bibinfo{author}{Rodionova, O.} \& \bibinfo{author}{Pomerantsev, A.}
\newblock \bibinfo{title}{Procrustes cross-validation—a bridge between cross-validation and independent validation sets}.
\newblock \emph{\bibinfo{journal}{Analytical Chemistry}} \textbf{\bibinfo{volume}{92}}, \bibinfo{pages}{11842--11850} (\bibinfo{year}{2020}).

\bibitem{ref_pcv2}
\bibinfo{author}{Kucheryavskiy, S.}, \bibinfo{author}{Rodionova, O.} \& \bibinfo{author}{Pomerantsev, A.}
\newblock \bibinfo{title}{Procrustes cross-validation of multivariate regression models}.
\newblock \emph{\bibinfo{journal}{Analytica Chimica Acta}} \textbf{\bibinfo{volume}{1255}}, \bibinfo{pages}{341096} (\bibinfo{year}{2023}).
\newblock \urlprefix\url{https://www.sciencedirect.com/science/article/pii/S0003267023003173}.

\bibitem{ref_simpls}
\bibinfo{author}{{de Jong}, S.}
\newblock \bibinfo{title}{Simpls: An alternative approach to partial least squares regression}.
\newblock \emph{\bibinfo{journal}{Chemometrics and Intelligent Laboratory Systems}} \textbf{\bibinfo{volume}{18}}, \bibinfo{pages}{251--263} (\bibinfo{year}{1993}).

\bibitem{ref_pytorch}
\bibinfo{author}{Paszke, A.} \emph{et~al.}
\newblock \bibinfo{title}{ in \textit{Pytorch: An imperative style, high-performance deep learning library}}  \bibinfo{pages}{8024--8035} (\bibinfo{publisher}{Curran Associates, Inc.}, \bibinfo{year}{2019}).
\newblock \urlprefix\url{http://papers.neurips.cc/paper/9015-pytorch-an-imperative-style-high-performance-deep-learning-library.pdf}.

\bibitem{ref_numpy}
\bibinfo{author}{Harris, C.~R.} \emph{et~al.}
\newblock \bibinfo{title}{Array programming with {NumPy}}.
\newblock \emph{\bibinfo{journal}{Nature}} \textbf{\bibinfo{volume}{585}}, \bibinfo{pages}{357--362} (\bibinfo{year}{2020}).
\newblock \urlprefix\url{https://doi.org/10.1038/s41586-020-2649-2}.

\bibitem{ref_tecator}
\bibinfo{author}{Borggaard, C.} \& \bibinfo{author}{Thodberg, H.~H.}
\newblock \bibinfo{title}{Optimal minimal neural interpretation of spectra}.
\newblock \emph{\bibinfo{journal}{Analytical Chemistry}} \textbf{\bibinfo{volume}{64}}, \bibinfo{pages}{545--551} (\bibinfo{year}{1992}).

\bibitem{ref_heart1}
\bibinfo{author}{Detrano, R.} \emph{et~al.}
\newblock \bibinfo{title}{International application of a new probability algorithm for the diagnosis of coronary artery disease}.
\newblock \emph{\bibinfo{journal}{The American Journal of Cardiology}} \textbf{\bibinfo{volume}{64}}, \bibinfo{pages}{304--310} (\bibinfo{year}{1989}).
\newblock \urlprefix\url{https://www.sciencedirect.com/science/article/pii/0002914989905249}.

\bibitem{ref_heart2}
\bibinfo{author}{Detrano, R.} \emph{et~al.}
\newblock \bibinfo{title}{Bayesian probability analysis: a prospective demonstration of its clinical utility in diagnosing coronary disease.}
\newblock \emph{\bibinfo{journal}{Circulation}} \textbf{\bibinfo{volume}{69}}, \bibinfo{pages}{541--547} (\bibinfo{year}{1984}).

\bibitem{ref_heart_data_repo}
\bibinfo{author}{Janosi, A.}, \bibinfo{author}{Steinbrunn, W.}, \bibinfo{author}{Pfisterer, M.} \& \bibinfo{author}{Detrano, R.}
\newblock \bibinfo{title}{{Heart Disease Dataset}} (\bibinfo{year}{1988}).

\end{thebibliography}
\end{document}